\documentclass[twoside]{article}

\usepackage{url}
\usepackage{graphicx}
\usepackage{natbib}
\usepackage{graphicx}
\usepackage{amsmath}
\usepackage{amssymb}
\usepackage{amsfonts}
\usepackage{amsthm}
\usepackage{multirow}
\usepackage{booktabs}
\usepackage{color}

\newcommand{\Rbb}{\mathbb{R}}

\newcommand{\Ib}{\mathbf{I}}

\DeclareMathOperator{\MAPE}{MAPE}

\DeclareMathOperator{\NMAE}{NMAE}

\begin{document}

\title{\large{\textbf{Electricity Demand Forecasting by Multi-Task Learning}}}
\author{Jean-Baptiste~Fiot, \qquad Francesco~Dinuzzo\\ {IBM Research - Ireland}}
\date{}

\maketitle

\begin{abstract}
We explore the application of kernel-based multi-task learning techniques to forecast the demand of electricity in multiple nodes of a distribution network. We show that recently developed output kernel learning techniques are particularly well suited to solve this problem, as they allow to flexibly model the complex seasonal effects that characterize electricity demand data, while learning and exploiting correlations between multiple demand profiles. We also demonstrate that kernels with a multiplicative structure yield superior predictive performance with respect to the widely adopted (generalized) additive models. Our study is based on residential and industrial smart meter data provided by the Irish Commission for Energy Regulation (CER).
\end{abstract}

\section{Introduction}
\label{sec:intro}

Electricity cannot be stored efficiently in large quantities, therefore it is critical to ensure that the amount generated at a given time is sufficient to meet the load plus the losses while not exceeding this amount significantly. Predictive methods for accurately forecasting the demand of electricity have thus become important tools that guide planning and operation of utility companies. While electric load forecasting is a well-established, several decades old research area in engineering, new modeling problems keep appearing as technological and legislative transformations affect the power industry. With the advent of smart grids and meters, larger and richer sources of data are becoming available, making it possible to build more sophisticated models that enable more accurate billing of electricity and dynamic pricing.

A variety of tools from time series analysis, statistics, and more recently machine learning, have been employed for electricity load forecasting. For an overview on the vast body of available literature on the subject, we refer the reader to the recent book by \cite{soliman2010electrical}. Classical techniques include linear and non-linear regression models estimated by means of variants of least squares fitting, and various types of ARMAX models expressing the forecast as a function of previously observed values of the load and possibly other weather or social variables. Techniques inspired by Artificial Intelligence research such as expert systems, fuzzy logic, and neural networks have also been applied to load forecasting. In particular, black-box models based on neural networks have been extensively analyzed, see the influential review by \cite{hippert2001neural}.

In recent years, Generalized Additive Models (GAM) \cite{hastie1990generalized} have established themselves as state of the art tools for electricity load forecasting \cite{fan2012short, ba2012adaptive, nedellec2013gefcom2012}, due to the existence of efficient and scalable training algorithms and the interpretability of the model, which allows to clearly visualize the effect of individual variables on the load by means of simple longitudinal plots. Meanwhile, kernel methods have been employed with great success in the last decade. Already back in 2001, a kernel-based Support Vector Regression (SVR) approach was employed to win a competition on electricity load forecasting \cite{chen2004load} organized by EUNITE (European Network on Intelligent Technologies for Smart Adaptive Systems). Later on, various types of kernel-based regularization methods and Support Vector Machines have been applied to predict the demand of electricity, see for instance \cite{espinoza2007electric, hong2009electric, elattar2010electric}.

Most research articles on electricity load forecasting focus on predicting a single time series representing the electricity load aggregated over a large number of nodes of the electricity network. Due to aggregation, such time series exhibit high regularity and are therefore significantly easier to forecast than load profiles at lower levels of the network. Nevertheless, making forecasts of the loads at lower levels is becoming increasingly feasible due to the availability of rich smart meter datasets, therefore the problem is attracting considerable interest in the industry.

Forecasting electricity demand at low levels of the network (such as the demand of an individual household) presents several challenges. First of all, it involves analyzing a much larger number of time series, calling for scalable techniques that can handle a very large amount of measurements. In addition, demand profiles at lower levels of the electricity network are much less regular and thus harder to predict. To tackle these challenges, recent works have investigated the use of clustering techniques for automatically aggregating multiple load time series, reporting improved predictive performance at aggregated level \cite{alzate2009identifying,alzate2013improved,humeau2013electricity}.

In this paper, we study the problem of electricity load forecasting at low network level, and we suggest to solve it by means of kernel-based multi-task learning techniques that can discover and take advantage of the relationships between multiple profiles. Kernel based multi-task learning has been studied in a variety of papers \cite{micchelli2004kernels,Evgeniou05,argyriou2007multi,bonilla2007multi} while, in recent years, the problem of learning and exploiting relationships between multiple tasks is a topic that is attracting considerable attention in the machine learning literature \cite{jacob2008clustered,Zhang08,Zhang10,Dinuzzo11, archambeau2011sparse,kang2011learning,Saha11,kumar12,Romera12, dinuzzo2013learning}.

Herein, we develop and compare a variety of kernel-based models for long-term electricity demand forecasting in multiple nodes, with the goal of identifying the best way to capture the complex seasonal effects that characterize such demand patterns. We design kernels specifically tailored to capture the seasonal effects present in electricity load data. By doing so, we expose the performance limits of the very popular additive models, showing that they are often outperformed by multiplicative kernel models. We formulate the problem of forecasting the demand in multiple nodes of the network as a multi-task learning problem, illustrating the usefulness of jointly learning and exploiting similarities between multiple load profiles. Finally, we show how recently developed multi-task learning techniques can be used to gain insights and interpretability on real demand data, while achieving state of the art predictive performance.  Our experimental analysis is based on data provided by the Irish Commission for Energy Regulation (CER).

\section{Electric load forecasting: goals and dataset structure}
\label{sec:lf}

Electric load forecasting aims at predicting the future load in one or multiple nodes of an electricity network. Depending on how far ahead in time the forecast is required, the corresponding estimation problem exhibits different characteristics, and influence decisions of significantly different nature. It is therefore common to classify forecasting problems in three categories: short-range forecasting (several minutes up to one week ahead), medium-range forecasting (up to 10 years ahead), and long-range forecasting (as far as several decades ahead), see \cite{soliman2010electrical} for a more comprehensive discussion.

Forecasting models are built starting from datasets containing one or multiple time series, each of them representing the load measurement at a specific location and level in the network, ranging from highly aggregated loads in the transmission network down to the distribution network and to the loads of individual users. Missing measurements and different sampling rates contribute to make these data noisy and challenging to analyze. Moreover, defective meters at a low level in the network are hard to detect, and faulty meters can report wrong measurements before being replaced.

\newcommand{\figW}{0.16}
\begin{figure}
\centering
\includegraphics[width=1.0\textwidth]{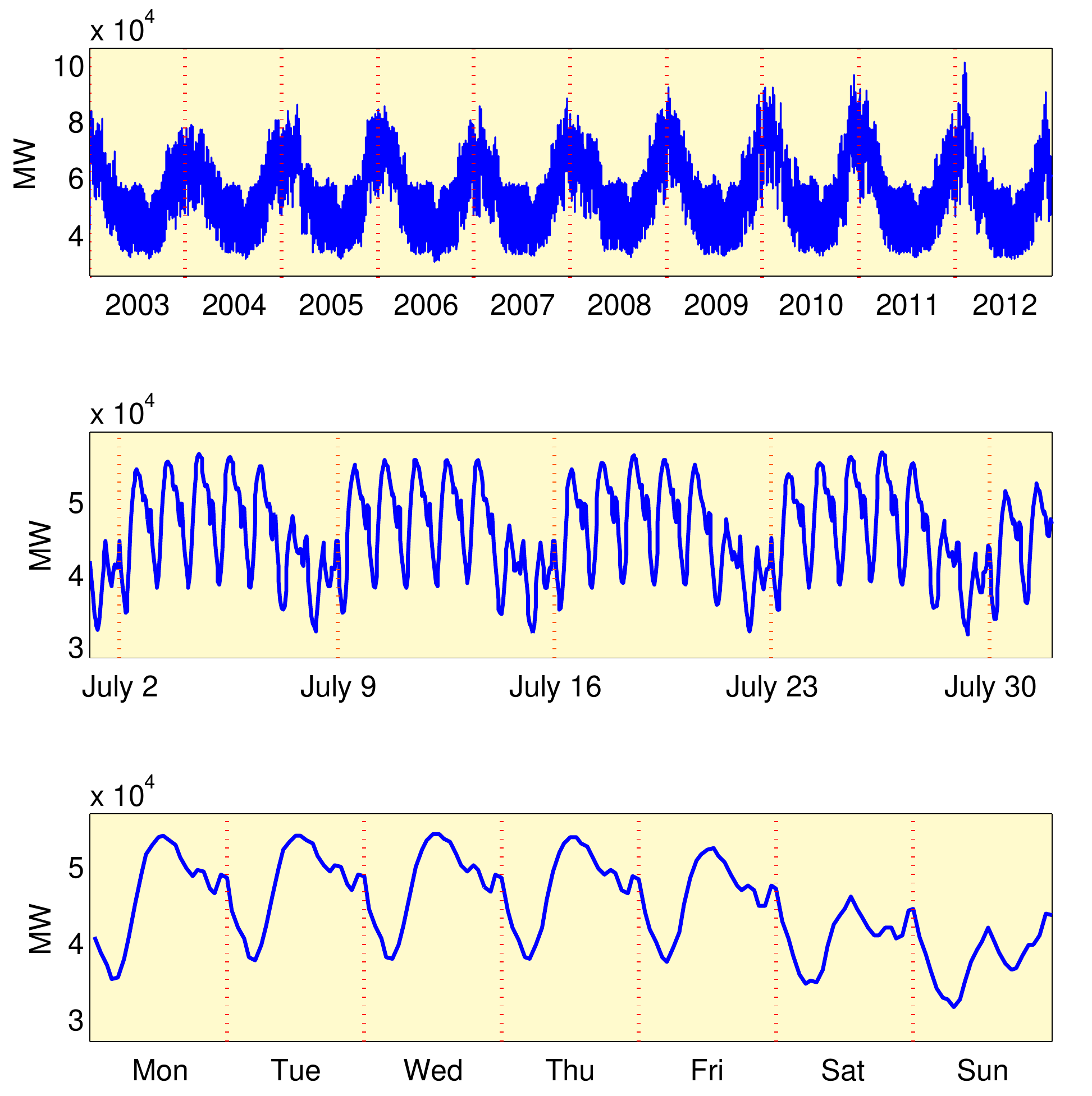}
\caption{Electric load data: yearly, weekly, and daily seasonal patterns can be observed in the top, middle, and bottom panels, respectively. }
\label{fig:patterns}
\end{figure}

\begin{figure}
	\centering
	\includegraphics[width=1.0\textwidth]{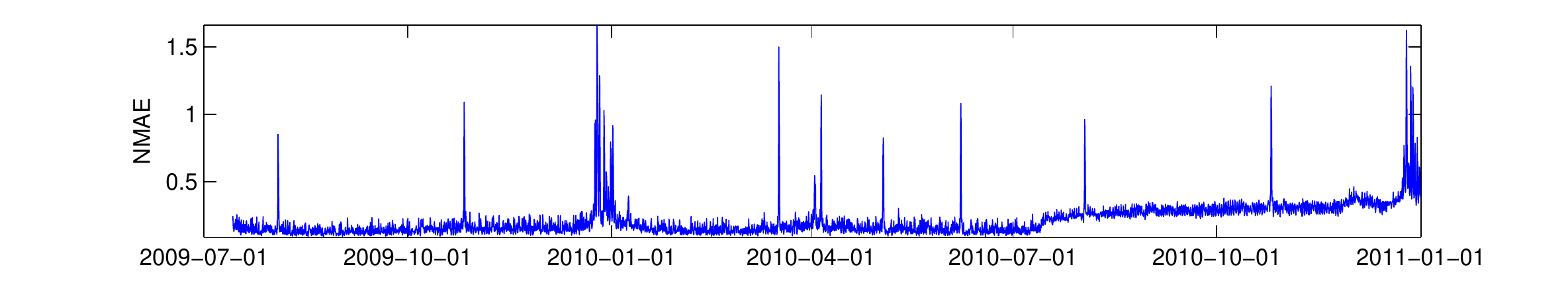}
	\caption{CER data (SME meters): without specific information about special events and holidays, the forecasting error can be particularly high in correspondence with those events. Uninformed models have difficulties to fit the demand even within the training period (until July 2010), let alone the test period (last six months). See section \ref{sec:LFbyMTL} for more details about CER data.}
	\label{fig:holidays}
\end{figure}

Being mostly driven by human activity, a variety of temporal patterns can be observed in the load data. The top panel of Fig.~\ref{fig:patterns} shows a typical profile for aggregated electricity load over several years (data source: French R\'eseau de Transport d'Electricit\'e\footnote{\url{http://clients.rte-france.com/lang/fr/visiteurs/vie/vie_stats_conso_inst.jsp}}), from which a clear yearly seasonal pattern can be observed, with higher demand in winter and lower demand in the summer. A closer look at this data also reveals typical weekly (Fig.~\ref{fig:patterns}, middle panel) and daily (Fig.~\ref{fig:patterns}, bottom panel) profiles. Correctly capturing these seasonal patterns is an crucial aspect of the problem, which can be dealt with by properly extracting and utilizing temporal and calendar features. The type of day of the week can be also taken into account: the bottom panel of Fig.~\ref{fig:patterns} shows a specific week where all days have a similar profile but a difference between week days and weekend can be clearly noticed. Forecasting is particularly challenging on public holidays, and different public holidays may exhibit significantly different load profiles. Fig. \ref{fig:holidays} illustrates the difficulty of fitting and forecasting the demand in correspondence of public holidays and special events: without including specific information in the model, the prediction error can be particularly high in corrispondence with such events.

A variety of additional features can be typically extracted from the data or obtained from other sources and utilized to forecast the electricity demand. For instance,  the electricity consumption is affected by weather conditions (particularly due to heating and air-conditioning), therefore variables such as temperature, humidity and irradiance are often taken into account by forecasting models. Economic indicators such as gross domestic product can be used to model trends in long-range scenarios. Finally, short term forecasting models are typically based on time series techniques, where auto-regressive lagged values of the load itself are incorporated in the model and used to track short-range trends and deviations from stationarity.

In summary, typical forecasts of the electricity demand may depend on a variety of features that include time and calendar variables, weather and economic conditions, previously observed values of the load, and information about the node of the network where the forecast is required. A general model that takes into account the previously discussed features takes the form
\begin{equation}\label{eq:general_model}
	\textup{Forecast} = f\left(\underbrace{t, d, c}_{\substack{\textup{Time / Calendar}\\ \textup{features}}},~\underbrace{y_l, u_l}_{\substack{\textup{Dynamic}\\\textup{features}}},~\underbrace{k, s_k}_{\textup{Node features}}\right),
\end{equation}
where the dependent variables are the following:
\begin{itemize}
  \itemsep0em
	\item $t \in [0, 24)$ is the time of day expressed in hours,
	\item $d \in \{1, 2, \ldots, 365, 366\}$ is the day of the year,
	\item $c$ is the type of day, e.g. Monday to Sunday, weekday/weekend, holiday,
    \item $y_l$ is a real vector containing lagged values of the measured electric load,
    \item $u_l$ is a real vector containing measurements of lagged values of exogenous variables other than the load (such as temperature),
    \item $k$ is the node ID in the electricity network, e.g. meter/customer/zone ID,
	\item $s_k$ is a vector of features describing the node type, e.g. customer/zone type.
\end{itemize}

In Section~\ref{sec:LFbyMTL}, we analyze one of the many possible multi-task learning problems that naturally appear within this framework, namely the problem of simultaneously predicting the load in each node of the network. This amount to disaggregate the overall dataset over the multiple smart meters and treat each node $k$ of the network as a different learning task. In the next section, we briefly recall the standard setup of multi-task regression and review the techniques that will be employed in Section~\ref{sec:LFbyMTL} to solve the problem sketched above.

\section{Kernel-based multi-task regression}
\label{sec:mtl}

In the following, we focus on multi-variate (multi-task) regression problems where the goal is to learn multiple functions $f_j: \mathcal{X} \rightarrow \mathbb{R}$ from multiple datasets of pairs $(x_{ij}, y_{ij}) \in \mathcal{X} \times \mathbb{R}$. Here, $\mathcal{X}$ is a set of input features, $m$ denotes the number of tasks, and $\ell_j$ the number of examples for the $j$-th task. Letting $f:\mathcal{X} \times \mathbb{R}^m$ denote the vector-valued function with components $f_j$, we are going to search $f$ by minimizing the following regularization functional
\[
R(f,\mathbf{L}) = \sum_{j=1}^{m}\sum_{i=1}^{\ell_j}\left(y_{ij}-f_j(x_{ij}))\right)^2 + \lambda \|f\|^2_{\mathcal{H}_{\mathbf{L}}}.
\]
where $\lambda > 0$ is a regularization parameter, and $\mathcal{H}_{\mathbf{L}}$ is a Reproducing Kernel Hilbert Space (RKHS) of vector-valued functions with (matrix-valued) kernel
\[
H(x_i,x_j) = K(x_i,x_j) \mathbf{L}.
\]
Here, $K:\mathcal{X} \times \mathcal{X} \rightarrow \mathbb{R}$ is a positive semidefinite kernel called \emph{input kernel}, and the square matrix $\mathbf{L} \in \Rbb^{m \times m}$ is the \emph{output kernel} matrix whose entries $\mathbf{L}_{jk}$ express the similarity between the tasks (output components) $j$ and $k$. In view of the \emph{representer theorem}, there exist functions $\hat{f}_j$ minimizing $R(f,\mathbf{L})$ in the form:
\begin{equation}\label{representer}
\hat{f}_j(x) =\sum_{k=1}^{m} \mathbf{L}_{jk} \sum_{i=1}^{\ell_k} c_{ik} K(x_{ik},x).
\end{equation}
We refer to \cite{Micchelli05} for more details about RKHS of vector-valued functions and the corresponding representer theorem.

\subsection{Fixing $\mathbf{L} = \Ib$ (independent kernel ridge regression)} \label{sec:krr}

Expression \eqref{representer} shows that inter-task transfer is possible only when off-diagonal elements of the output kernel matrix are different from zero. Indeed, by choosing $\mathbf{L}$ equal to the identity matrix all the tasks are learned independently by solving a standard kernel regularized least squares problem
\begin{equation}\label{eq:independent}
\hat{f}_j = \arg \min_{f_j \in \mathcal{H}} \left(\sum_{i=1}^{\ell_j}\left(y_{ij}-f_j(x_{ij})\right)^2+ \lambda \|f_j\|^2_{\mathcal{H}}\right),
\end{equation}
\noindent where $\mathcal{H}$ is the RKHS of scalar functions with kernel $K$. 
This single-task baseline is referred to as \emph{independent} kernel ridge regression.

\subsection{Learning $\mathbf{L}$ (output kernel learning)} \label{sec:okl}

In this subsection, we review a kernel-based multi-task regression approach called low-rank Output Kernel Learning (OKL), recently developed in \cite{dinuzzo2013learning}. In such approach, the functions $f_j$ and the output kernel $\mathbf{L}$ are jointly optimized by solving the following problem
\begin{equation}
  \label{EQ02} \min_{\mathbf{L} \in \mathbb{S}^{m,p}_+} \quad \min_{f \in \mathcal{H}_{\mathbf{L}}} \quad R(f,\mathbf{L}) + \lambda \textup{tr}(\mathbf{L}),
\end{equation}
\noindent where $\mathbb{S}^{m,p}_+$ is the cone of positive semidefinite matrices with rank less than or equal to $p$. Instead of imposing a low-rank constraint or regularizing the trace of the output kernel, other type of regularizers could be tried, see e.g. \cite{Dinuzzo11, dinuzzo2014output, ciliberto15}. The low-rank approach has the advantage of allowing us to tightly control the memory required to store the models.

The representer theorem \eqref{representer} still applies to the inner minimization problem of \eqref{EQ02}. By plugging the expression \eqref{representer} into \eqref{EQ02}, one obtains a functional that is convex quadratic with respect to both the coefficients $c_{ik}$ and $\mathbf{L}$. Although the resulting problem is not jointly convex, the alternating minimization procedure described in \cite{dinuzzo2013learning} can be applied to obtain a minimizer. An important aspect of the method is that, by selecting the rank parameter $p$, is it possible to control the overall number of parameters of the model, as well as the memory requirements and the computation time to obtain a solution. More specifically, letting $\mathcal{A} = \cup_{j}\cup_i\{x_{ij}\}$, one can show that the solution \eqref{EQ02} can be rewritten as
\begin{equation}\label{eq:modes}
\hat{f}_j(x) =\sum_{k=1}^{p} b_{jk} g_k(x), \quad g_k(x) = \sum_{i=1}^{\ell} a_{ik} K(x_{i},x),
\end{equation}
where $\ell = \#\mathcal{A}$, $x_i \in \mathcal{A}$, $i =1, \ldots, \ell$, and the coefficients $b_{jk}$ form a low-rank factor of $\mathbf{L}$. It is therefore sufficient to store and optimize $(\ell + m) p$ parameters, which can be much smaller than $\sum_{j=1}^{m} \ell_j$.

\section{Electricity demand forecasting in multiple nodes by multi-task learning}
\label{sec:LFbyMTL}
In this section, we focus on predicting the demand of electricity in multiple nodes of an electricity network, a multi-task learning problem where each task corresponds to one of the measured smart meters in the network.

\renewcommand{\arraystretch}{1.2}
\begin{table}
	\caption{Number of meters and sparsity for each customer group in the Irish CER dataset}
	\label{tab:groups}
	\centering
	\begin{tabular}{@{}lrr@{}}
		\toprule
		\textbf{Customer group} & \textbf{Meters} & \textbf{Sparsity} \\ \midrule
		Residential & 4225 & 0.028\% \\ 
		Industrial (SME) & 485 & 0.035\% \\
		Others & 1723 & 17\% \\ \bottomrule
	\end{tabular}
\end{table}

To analyze our long-term forecasting approach, we adopt data provided by the the Irish Commission for Energy Regulation (CER) \footnote{\url{http://www.ucd.ie/issda/data/commissionforenergyregulationcer/}}, containing electric load measurements from 6435 smart meters, half-hourly sampled from July 14, 2009 to December 31, 2010 (536 days). These meters include residential customers and small-to-medium industrial sites. We consider a mid-term test scenario where the goal is to forecast the load in multiple nodes over a time horizon of 171 days, using one year of measurements to build the model. Due to the long forecasting horizon, dynamic features are not available and are therefore dropped from the general model in Eq.~\eqref{eq:general_model}. Such model does not rely on recent measurements of the load, therefore it is able to make predictions over an arbitrarily long horizon. The load forecast for the $k$-th node is thus simply given by $\hat{y} = f_k(t, d, c)$, where $f_k$ are the multiple functions to be learned, taking into account time and calendar features.

From the original CER dataset, several pre-processing steps were performed. The day of the year and time of the day were extracted from the five-digit timestamps. In this dataset, the time of day is an non-zero integer indexing the number of half-hours, and therefore it should be normally in the set $\{1, 2, \dots, 48\}$. Two meters containing time of days higher than $50$ half-hours were discarded, as it was unclear how to interpret these measurements. The dataset also contains days with 46 and 50 measurements and time of days up to 50. These inconsistencies are caused by the start and the end of daylight saving time (DST) and are easily fixable. When DST starts in Ireland \footnote{\url{http://www.timeanddate.com/time/change/ireland/dublin}}, the 1AM to 2AM hour get skipped, and half-hourly time of day indices should be $\{1, 2, 5, 6, \dots, 48\}$. When DST starts in Ireland, the 1AM to 2AM hour ``happens twice'', and half-hourly time of day indices should be $\{1, 2, 3, 4, 3, 4, 5, 6, \dots, 48\}$, instead of $\{1, 2, \dots, 50\}$ as found in the dataset. We then downsampled each time-series from half-hourly sampling to 3-hour sampling, by averaging available measurements for each time slot of 3 hours ([12AM, 3AM), [3AM, 6AM), etc) and a total of 8 measurements per day. Our final dataset contains $m=6433$ smart meters sampled over $\ell=4288$ time slots. Characteristics of such pre-processed dataset are summarized in Tab.~\ref{tab:groups}. One year ($2920$ downsampled observations) was used for training and validation, and the remaining $1368$ observations were used for testing. In order to perform tuning of the regularization parameter, we extracted a validation set containing a subset of the original non-test data, obtained by randomly choosing $20\%$ time samples, equal for all the meters.

Forecasting performance can be evaluated for each time slot $i=1, \ldots, \ell$ and any arbitrary group of meters $\mathcal{G}$. For this purpose, let $\mathcal{G}_i$ denote the subset of $\mathcal{G}$ for which measurements are available in the $i$-th time slot. We define two different metrics, namely the aggregated mean absolute percentage error ($\MAPE$) and the normalized mean absolute error ($\NMAE$):
\begin{align*}
	\MAPE(i,\mathcal{G}) & =   100 \frac{ \left|\sum_{j \in \mathcal{G}_i} y_{ij} - \sum_{j \in \mathcal{G}_i} f_j(t_i,d_i,c_i)\right|}{\sum_{j \in \mathcal{G}_i} y_{ij} }, \\
	\NMAE(i,\mathcal{G}) & =   \frac{\sum_{j \in \mathcal{G}_i} \left|y_{ij} - f_j(t_i,d_i,c_i)\right|}{\sum_{j \in \mathcal{G}_i}  y_{ij} }. 
\end{align*}

$\MAPE(i,\mathcal{G})$ measures the relative absolute percentage error incurred when forecasting the aggregated demand using the sum of the forecasts. On the other hand, $\NMAE(i,\mathcal{G})$ is the sum of the forecasting errors over individual tasks, relative to the naive baseline of predicting $f_j(t_i,d_i,c_i)=0$ for all $i,j$. Since the demand values $y_{ij}$ are always non-negative, the two metrics are undefined only for those groups on meters for which the cumulative demand in the $i$-th time slot is identically zero, or no measurements are available for any of the meters. We compute the average and standard deviation of these two metrics over all the time slots in the test period.

One advantage of the long-term forecasting approach is that it allows to naturally incorporate and handle time series with missing observations, without resorting to inputing techniques or discarding data. In the following, we analyze a variety of kernel based models to solve this multi-task regression problem. First of all we introduce kernels based on the time/calendar features of model~\eqref{eq:general_model}, 
\begin{align*}
	K^t(t_1,t_2) & = \exp\left( - h_T(|t_1-t_2|) / \sigma_t \right),\\
	K^d(d_1,d_2) & = \exp\left( - h_D(|d_1-d_2|) / \sigma_d \right),\\
    K^{c}(c_1, c_2) & = \begin{cases} 1 &\mbox{if } c_1 = c_2 \\ 
    0 & \mbox{if }  c_1 \neq c_2. \end{cases},
\end{align*}
where $h_P(x) = \min \{x, P-x\}$ is a change of variable that yields $P$-periodic kernels over the square $[0, P]^2$. By observing that the Fourier transform of $\exp(-|x|)$ is non-negative, it can be easily shown that periodized kernels such as $K^t$ and $K^d$ are positive semidefinite, see \cite{Scholkopf01b}. In our experiment, $\sigma_t$ and $\sigma_d$ were respectively set to 4 hours and 120 days. In order to define $K((t_1, d_1, c_1), (t_2, d_2, c_2))$, we combine these three kernels to define a variety of models:
\begin{itemize}
  \itemsep0em
	\item Additive Models
	  \begin{align}
	     &K^{t}(t_1, t_2) + K^{d}(d_1,d_2)\,, \label{eq:AM1} \\
	     &K^{t}(t_1, t_2) + K^{d}(d_1,d_2) + K^{c}(c_1,c_2)\,, \label{eq:AM2}
	  \end{align}
	\item Semi-Additive Models
	  \begin{align}
	    &K^{d}(d_1,d_2) + K^{t}(t_1, t_2)\cdot K^{c}(c_1,c_2)\,, \label{eq:SAM1}\\
	    &\left(K^{t}(t_1, t_2) + K^{d}(d_1,d_2) \right) \cdot K^{c}(c_1,c_2)\,, \label{eq:SAM2}
    \end{align}
  \item Multiplicative Models
    \begin{align}
	    & K^{t}(t_1, t_2) \cdot K^{d}(d_1,d_2)\,, \label{eq:MM1} \\ 
	    & K^{t}(t_1, t_2) \cdot K^{d}(d_1,d_2) \cdot K^{c}(c_1,c_2)\,,
  	\label{eq:MM2} 
  \end{align}
\end{itemize}

First of all, we have trained independent kernel ridge regression models (see  Sec.~\ref{sec:krr}) for each measured smart meter using all the kernels from \eqref{eq:AM1} to \eqref{eq:MM2}.  We compare these models against a multi-task learning approach that simultaneously performs estimates for all the meters, and also allows us to exploit the available meter grouping information in the dataset. Specifically, we have trained two separate multi-task output kernel learning (OKL) models (see Sec. \ref{sec:okl}): the first is trained over all the residential meters (the union of meters labeled ``residential'' and ``others''), and the second over industrial (SME) meters. The maximum rank constraint for the first model was set to $p=200$ to obtain a compact model that fits into memory, while the OKL model for SME meters was trained with full rank $p = 485$. We refer the reader to \cite{dinuzzo2013learning} for a discussion on the effect of this parameter. Both OKL models utilize the multiplicative input kernel \eqref{eq:MM2}, as it proves to be the most effective at capturing the seasonal effects. 

\begin{figure}
	\centering
	\includegraphics[width=1\textwidth]{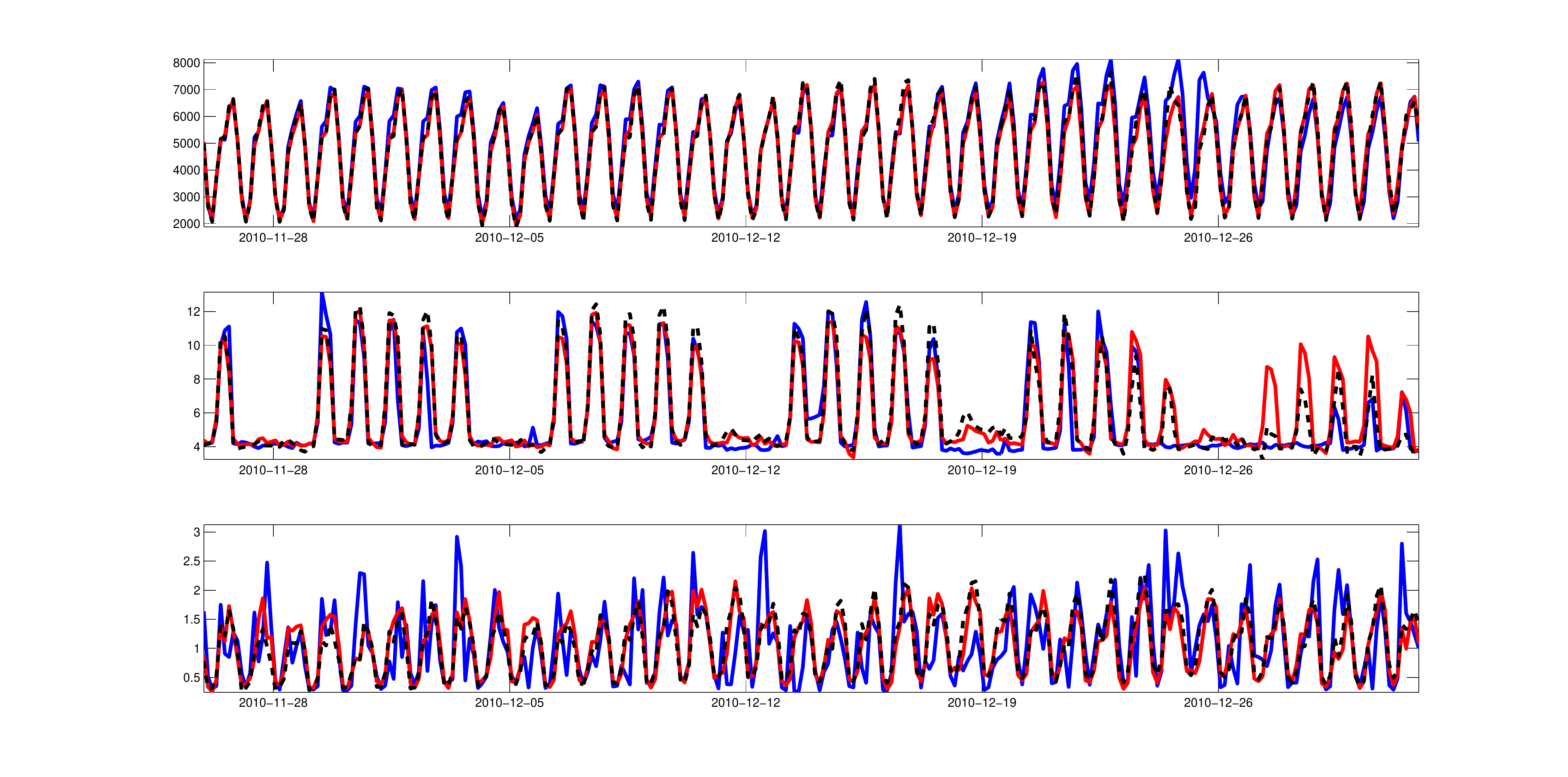}	
	\caption{CER data: measured load (blue curves) and corresponding forecast with independent (red curve) and multi-task (black dashed curve) for the aggregated demand (top panel), a single SME meter (mid panel), and a single residential meter (bottom panel). Measurements are shown over 5 weeks of the test period. All forecasts are obtained using a multiplicative kernel model \eqref{eq:MM2}.}
	\label{fig:predictions}
\end{figure}

Figure~\ref{fig:predictions} illustrates the challenges of long-term forecasting at low network level versus forecasting aggregated demands. We analyze the measured load and the corresponding forecast over a window of 5 weeks within the test period. In the top panel, one can see the aggregated load and the corresponding forecast obtained as the sum of all disaggregated forecasts obtained using model~\eqref{eq:MM2}. The kernel-based forecasts are rather accurate overall, only slighly estimating the total load during the Christmas week, a particularly problematic period to predict. In the middle panel, the measured load for a single SME meter is compared with the corresponding forecast. The varying demand profiles of different days of the week are captured rather well by the model. Again, there is a larger error over the Christmas week, caused by a sudden drop of the demand to a low baseline value (probably due to interruption of business activities followed by a slow resumption in the subsequent days). This leads the models to over-estimate the load, though the model learned with a multi-task approach is less affected.  Finally, the bottom panel shows the electricity demand of a residential customer, characterized by rapid variations with sharp consumption peaks and irregular patterns, that make the forecast even more difficult.

\renewcommand{\arraystretch}{1.2}

\begin{table*}
  \caption{Overall accuracy of the different long-term forecasting methods on the CER dataset}
  \label{tab:CERresults_overall}
  \centering
  \begin{tabular}{@{}llrrrrr@{}}
	  \hline{}
    \multirow{2}{*}{Method} & \multirow{2}{*}{Kernel} & \multicolumn{2}{c}{NMAE} & \phantom{abc} & \multicolumn{2}{c}{MAPE} \\ \cline{3-4} \cline{6-7} 
     && Mean & Std & & Mean & Std   \\ \midrule
     Additive Models & $K^d+K^t$ & 0.4829 & 0.0834 && 8.0384 & 7.2644
     \\
                     & $K^d+K^t+K^c$ & 0.4897 & 0.0846 && 8.0435 & 6.9948
                     \\ \\
     Semi-Additive Models & $K^d+K^t \cdot K^c$ & 0.4546 & 0.0682 && 7.1413 & 6.6078
     \\
                     & $\left(K^d+K^t\right)\cdot K^c$ & 0.4507 & 0.0649 && 7.2921 & 6.5868
                     \\ \\
     Multiplicative Models & $K^d \cdot K^t$ & 0.4663 &  0.0771 && 5.5284 & 5.4252
     \\
                     & $K^d\cdot K^t \cdot K^c$ & 0.4237 & 0.0528 && 4.2917 & 4.3055
                     \\ \\
	   Multi-Task OKL & $K^d\cdot K^t \cdot K^c$ & \textbf{0.4226} & \textbf{0.0487} && \textbf{4.0222} &
\textbf{4.0541}
	   \\ \bottomrule
  \end{tabular}
\end{table*}

\begin{table*}
  \caption{Accuracy of the different long-term forecasting methods on the residential customer group}
  \label{tab:CERresults_res}
  \centering
  \begin{tabular}{@{}llrrrrr@{}}
	  \hline{}
    \multirow{2}{*}{Method} & \multirow{2}{*}{Kernel} & \multicolumn{2}{c}{NMAE} & \phantom{abc} & \multicolumn{2}{c}{MAPE} \\ \cline{3-4} \cline{6-7} 
     && Mean & Std & & Mean & Std  \\ \midrule
     Additive Models & $K^d+K^t$ & 0.5114 & 0.0929 && 13.1083 & 9.9519 \\
                     & $K^d+K^t+K^c$ & 0.5157 & 0.0941 && 13.0127 & 10.2067 \\ \\
     Semi-Additive Models & $K^d+K^t \cdot K^c$ & 0.5005 & 0.0900 && 10.7139 & 9.9966 \\
                     & $\left(K^d+K^t\right)\cdot K^c$ & 0.4977 & 0.0846 && 10.8144 & 9.6131 \\ \\
     Multiplicative Models & $K^d \cdot K^t$ & 0.5058 & 0.0844 && 8.4289 & 7.1833 \\
                     & $K^d\cdot K^t \cdot K^c$ & 0.4776 & 0.0721 && 5.2692 & 5.6305 \\ \\
	   Multi-Task OKL & $K^d\cdot K^t \cdot K^c$ & \textbf{0.4711} & \textbf{0.0716} && \textbf{4.9166} &
\textbf{5.2588} \\ \bottomrule
  \end{tabular}
\end{table*}

\begin{table*}
  \caption{Accuracy of the different long-term forecasting methods on the SME customer group}
  \label{tab:CERresults_sme}
  \centering
  \begin{tabular}{@{}llrrrrr@{}}
	  \hline{}
    \multirow{2}{*}{Method} & \multirow{2}{*}{Kernel} & \multicolumn{2}{c}{NMAE} & \phantom{abc} & \multicolumn{2}{c}{MAPE} \\ \cline{3-4} \cline{6-7} 
     && Mean & Std & & Mean & Std  \\ \midrule
     Additive Models & $K^d+K^t$ & 0.4517 & 0.2204 && 15.0748 &
23.1005 \\
                     & $K^d+K^t+K^c$ & 0.4590 & 0.1769 && 16.6668 &
17.0665 \\ \\
     Semi-Additive Models & $K^d+K^t \cdot K^c$ & 0.3704 & 0.1238 && 8.1880 &
10.8081 \\
                     & $\left(K^d+K^t\right)\cdot K^c$ & 0.3646 & 0.1305 && 8.1422 &
11.7269 \\ \\
     Multiplicative Models & $K^d \cdot K^t$ & 0.4006 & 0.1923 && 12.5359 &
20.4140   \\
                     & $K^d\cdot K^t \cdot K^c$ & \textbf{0.3127} & 0.1105 && 5.9289 &
10.3842 \\ \\
	   Multi-Task OKL & $K^d\cdot K^t \cdot K^c$ & 0.3194 & \textbf{0.0725} && \textbf{5.2940} &
\textbf{5.8227} \\ \bottomrule
  \end{tabular}
\end{table*}

\begin{table*}
  \caption{Accuracy of the different long-term forecasting methods for the smart meters labeled as ``others''}
  \label{tab:CERresults_others}
  \centering
  \begin{tabular}{@{}llrrrrr@{}}
	  \hline{}
    \multirow{2}{*}{Method} & \multirow{2}{*}{Kernel} & \multicolumn{2}{c}{NMAE} & \phantom{abc} & \multicolumn{2}{c}{MAPE} \\ \cline{3-4} \cline{6-7} 
     && Mean & Std & & Mean & Std  \\ \midrule
     Additive Models & $K^d+K^t$ & 0.4948 & 0.0779 && 7.4672 &
6.9361 \\
                     & $K^d+K^t+K^c$ & 0.5001 & 0.0804 && 7.4141 & 
6.4660 \\ \\
     Semi-Additive Models & $K^d+K^t \cdot K^c$ & 0.4659 & 0.0638 && 6.4749 &
5.9244 \\
                     & $\left(K^d+K^t\right)\cdot K^c$ & 0.4601 & 0.0605 && 6.6273 &
5.9189 \\ \\
     Multiplicative Models & $K^d \cdot K^t$ & 0.4801 & 0.0710 && 5.4504 &
5.5997   \\
                     & $K^d\cdot K^t \cdot K^c$ & 0.4370 & 0.0517 && 4.0279 &
4.2522 \\ \\
	   Multi-Task OKL & $K^d\cdot K^t \cdot K^c$ & \textbf{0.4361} & 0.0490 && \textbf{3.7450} &
\textbf{3.9503} \\ \bottomrule
  \end{tabular}
\end{table*}

Table~\ref{tab:CERresults_overall} reports the performance of all methods over the full set of 6433 smart meters, while Tables~\ref{tab:CERresults_res}, \ref{tab:CERresults_sme} and \ref{tab:CERresults_others} report disaggregated performance measures over each group from Tab.~\ref{tab:groups}. We start by analyzing the performance of the additive models, which are probably the most widely adopted in the literature. By comparing the performance of models~\eqref{eq:AM1} and~\eqref{eq:AM2}, we can observe that adding a constant bias specific to the type of the day of week (kernel $K^c$) does not necessarily improve the accuracy of the model. The overall $\NMAE$ and $\MAPE$ are in fact higher for model~\eqref{eq:AM2}, see Tab.~\ref{tab:CERresults_overall}. Semi-additive models where the type of day of the week is utilized to switch between different profiles yields a significant improvement in performance. The two semi-additive models~\eqref{eq:SAM1} and~\eqref{eq:SAM2} achieve similar performance over the groups residential and others (see respectively Tab.~\ref{tab:CERresults_res} and Tab.~\ref{tab:CERresults_others}). However, for the SME customer group, model~\eqref{eq:SAM2} is better in terms of both $\NMAE$ and $\MAPE$ (see Tab.~\ref{tab:CERresults_sme}). In previous works such as \cite{ba2012adaptive}, semi-additive models of the form \eqref{eq:SAM1} have been proposed to switch between different daily patterns, depending on the type of day. Interestingly, our results show that in certain situations, such as when modeling industrial customers, it is even better to switch the overall sum of the daily pattern and the yearly pattern. We took a step even further by utilizing fully multiplicative models \eqref{eq:MM1} and~\eqref{eq:MM2}. The multiplicative model \eqref{eq:MM1} pools over different days of the week, while \eqref{eq:MM2} learns independent models for each day. While the former is not always better than the semi-additive models, the latter significantly outperforms them. We can conclude that a multiplicative kernel structure~\eqref{eq:MM2} is the best at capturing yearly, weekly and daily seasonal effect, both overall and for each customer group. Such conclusion is aligned with recent results presented in \cite{guerini15}, where tensor product basis functions were utilized to capture weekly and yearly seasonalities in the simpler context of load forecasting for a single highly aggregated time series. A further performance improvement can be obtained by utilizing a multi-task learning approach, where correlation between electricity demand behavior of multiple customers is learned and exploited. As the Tables show, the multi-task OKL approach provides the lowest mean $\NMAE$ over all the meters performance, residential and others, and second lowest mean $\NMAE$ for SME (only $2\%$ higher than the lowest for this group). The multi-task learning approach also provides the lowest mean aggregated $\MAPE$, overall and for each customer groups. Finally, the multi-task approach is more robust, as the temporal standard deviation is the lowest for both $\NMAE$ and $\MAPE$. Again, such robustness can be observed overall the customers as well as for each customer group. In particular, it is worth mentioning a $44\%$ improvement of the standard deviation of the $\MAPE$ for SME meters, compared to the best single task model that uses the multiplicative model~\eqref{eq:MM2}. 

In addition to improving forecasting accuracy, the low-rank multi-task learning model is significantly more compact in terms of number of parameters. For all the single-task methods (with additive, semi-additive and multiplicative kernels), the number of parameters is equal to the overall number of training observations $\sum_{j=1}^m \ell_j$. In our experiment, this amounts to about 13~million parameters (precisely $12 785 524$ parameters). The low-rank output kernel learning method models each prediction function $\hat{f}_j$ as a linear combination of $p$ latent functions, shared by all tasks  (see Sec.~\ref{sec:okl}). These $p$ functions $g_k$ can be seen as typical load profiles. As a consequence, only $(\ell + m)p$ parameters are required to learn the prediction functions for all smart meters. In our experiment, this gave a total of about 3~million parameters (precisely $3 016 310$) thus producing a model that is about $4.24$ times more compact, in addition to being more accurate.

\section{Discussion and conclusions}
\label{sec:ccl}

Our analysis shows that kernel-based multi-task learning is effective for the resolution of electric load forecasting problems. Focusing on the challenging problem of forecasting the electric load of individual customers, we designed kernels that take into account relevant multiple seasonality patterns. We demonstrated the clear benefits of multiplicative kernel models over additive or semi-additive models. Our results suggest a new modeling direction, as opposed to the (generalized) additive models, widely employed in the energy community.
We illustrated further performance gain made possible by using a multi-task learning approach over a large number of single-tasks baselines. While recent studies reported $\MAPE$ around $3\%$ for the short term forecasting of an aggregated signal of a few thousands of smart meters e.g. \cite{alzate2013improved}, our method achieves a $\MAPE$ of $4\%$ on a long term forecasting scenario, which is a much harder problem as auto-regressive terms are not available. 

The ideas and results presented in this paper open a wide range of considerations. First of all, they suggest that electricity demand data can be used as natural test benchmarks for multi-task learning methods. In addition, these problems motivate developing new techniques that allow to incorporate more complex task relationships structures taking into account, for instance, topological and physical constraints from the electricity network. The development of online methods that can automatically discover relationships between multiple tasks seems to be particularly important for short-term load forecasting scenarios. Finally, combining online multi-task learning methods with topological network constraints would allow to start tackling very complex scenarios such as forecasting on a full electricity network with dynamic reconfigurations.

\small
\bibliographystyle{unsrt}

\end{document}